\documentclass{article}

\usepackage{PRIMEarxiv}

\usepackage[utf8]{inputenc} 
\usepackage[T1]{fontenc}    
\usepackage{hyperref}       
\usepackage{url}            
\usepackage{booktabs}       
\usepackage{amsfonts}       
\usepackage{nicefrac}       
\usepackage{microtype}      
\usepackage{lipsum}
\usepackage{fancyhdr}       
\usepackage{graphicx}       

\usepackage{graphicx}%
\usepackage{multirow}%
\usepackage{amsmath,amssymb,amsfonts}%
\usepackage{amsthm}%
\usepackage{mathrsfs}%
\usepackage[title]{appendix}%
\usepackage{xcolor}%
\usepackage{textcomp}%
\usepackage{manyfoot}%
\usepackage{booktabs}%
\usepackage{algorithm}%
\usepackage{algorithmicx}%
\usepackage{algpseudocode}%
\usepackage{listings}%
\usepackage{array}
\usepackage{nicematrix}
\usepackage{times}
\usepackage{dsfont}
\usepackage{epsfig}
\usepackage{graphicx}
\usepackage{booktabs}
\usepackage{amsmath}
\usepackage{enumitem}
\usepackage[dvipsnames]{xcolor}
\usepackage{amssymb}
\usepackage[algo2e,ruled,vlined]{algorithm2e}
\usepackage{algorithm}
\usepackage{subcaption}
\usepackage{balance}
\usepackage{comment}
\usepackage{xcolor}
\usepackage{bbm}
\usepackage[nameinlink]{cleveref}
\usepackage{placeins}
\crefname{figure}{Fig.}{Figs.}
\Crefname{figure}{Fig.}{Figs.}

\newcounter{sitext}
\newcommand{\sitext}[2]{%
  \refstepcounter{sitext}%
  \subsubsection*{\thesitext. #1}%
  \phantomsection
  \label{#2}%
}

\newcommand{\kj}[1]{\textcolor{blue}{#1}}

\graphicspath{{media/}}     

\title{Mapping Reduced Accessibility to WASH Facilities \\
in Rohingya Refugee Camps With Sub-Meter Imagery
}

\author{
  Kyeongjin Ahn\textsuperscript{1,2,\dag} \ \ \ 
  YongHun Suh\textsuperscript{2,\dag} \ \ \ 
  Sungwon Han\textsuperscript{3,}\thanks{Participated in this work during an internship at MPI-SP.} \\
  \textbf{Jeasurk Yang\textsuperscript{2,\S}} \ \ \  
  \textbf{Hannes Taubenb{\"o}ck\textsuperscript{4,5,\S}} \ \ \  
  \textbf{Meeyoung Cha\textsuperscript{1,2,\S}} \\
  \\
  \textsuperscript{1}School of Computing, Korea Advanced Institute of Science and Technology (KAIST), \\ Daejeon, Republic of Korea \\ \\
  \textsuperscript{2}Data Science for Humanity Group, Max Planck Institute for Security and Privacy (MPI-SP), \\ Bochum, Germany \\ \\
  \textsuperscript{3}Meta, \\
  California, United States \\ \\
  \textsuperscript{4}German Aerospace Center (DLR), Earth Observation Center (EOC), \\ We{\ss}ling, Germany \\ \\
  \textsuperscript{5}Institute of Geography and Geology, Earth Observation Research Cluster (EORC), W{\"u}rzburg University, \\ W{\"u}rzburg, Germany \\ \\
  \textsuperscript{\dag}These authors contributed equally as first authors: \kj{kyeongjin.ahn@kaist.ac.kr}; \kj{yong-hun.suh@mpi-sp.org}. \\
  \textsuperscript{\S}Corresponding authors: \kj{jeasurk.yang@mpi-sp.org}; \kj{hannes.taubenboeck@dlr.de}; \kj{mia.cha@mpi-sp.org}.
}

\begin{document}
\maketitle

\begin{abstract}
Lack of access to Water, Sanitation, and Hygiene (WASH) services is a major public health concern in refugee camps, where extreme crowding accelerates the spread of communicable diseases. The Rohingya settlements in Cox’s Bazar, Bangladesh, exemplify these conditions, with large populations living under severe spatial constraints. We develop a semi-supervised segmentation framework using the Segment Anything Model (SAM) to map shelters from multi-temporal sub-meter remote sensing imagery (\textit{2017}–\textit{2025}), improving detection in complex camp environments by 4.9$\%$ in F1-score over strong baselines. The detected shelter maps show that shelter expansion stabilized after \textit{2020}, whereas continued population growth reduced per capita living space by $\sim$14$\%$ between \textit{2020} and \textit{2025}. WASH accessibility, measured with an enhanced network-based two-step floating catchment area (2SFCA) method, declined from \textit{2022} to \textit{2025}, increasing facility loads and exceeding global benchmarks. Gender-disaggregated scenarios that incorporate safety penalty further reveal pronounced inequities, with female accessibility $\sim$27$\%$ lower than male. Together, these results demonstrate that remote sensing–driven AI diagnostics can generate equity-focused evidence to prioritize WASH investments and mitigate health risks in protracted displacement settings (Code is available at \url{https://github.com/DS4H-GIS/Refugee_WASH_Accessibility}.).
\end{abstract}


\clearpage
\section{Introduction}
Forced displacement poses a significant global challenge, driven by armed conflicts, environmental hazards, and systemic persecution, which impose considerable burdens on resources and service provision in critical sectors such as education, livelihoods, and health \cite{iDMC25_IDP}. As of the end of \textit{2024}, the United Nations High Commissioner for Refugees (UNHCR) reported approximately 123 million forcibly displaced individuals, encompassing refugees and internally displaced persons (IDPs), thereby emphasizing the necessity for sustained international assistance \cite{taylor2016economic,UNHCR25_IDP,bertassello2023food,scharbert2024psychological,adema2025refugees}.

The Rohingya situation stands as a paradigmatic case of forced displacement. In \textit{2017}, violence in Myanmar’s Rakhine State compelled over 1 million people to flee to Bangladesh’s Cox’s Bazar District, leading to the establishment of extensive refugee camps \cite{RRR25rohingya,faye2021forced}. Designated as Forcibly Displaced Myanmar Nationals (FDMNs), this population resides in 33 camps spanning a total area of just 24 $km^2$. By \textit{May 2025}, these camps accommodated more than 1.1 million people, with an average density exceeding 45,000 individuals per $km^2$ \cite{UNHCR25site}—equivalent to population of Cologne, Germany’s fourth-largest city, but confined to less than one-$16^{\mathrm{th}}$ of its area.

Such high density amplifies the sectoral strains noted earlier—especially in health—by increasing the spread of communicable diseases and making public health the foremost priority for population survival \cite{Butt25WASH,shapna2023water}. Essential infrastructure, such as Water, Sanitation, and Hygiene (WASH) facilities, is crucial for mitigating these risks, yet persistent deficiencies exacerbate vulnerabilities across the population \cite{toma2018rohingya}. Despite the pressing need for better understanding, a comprehensive evaluation of camp-wide living conditions remains limited, as field surveys provide only localized data on service access \cite{akhter2020drinking,akter2021investigating} and lack the spatial granularity to uncover broader patterns of deficiencies.

\begin{figure*}[t!]
\centering
\includegraphics[width=16cm]{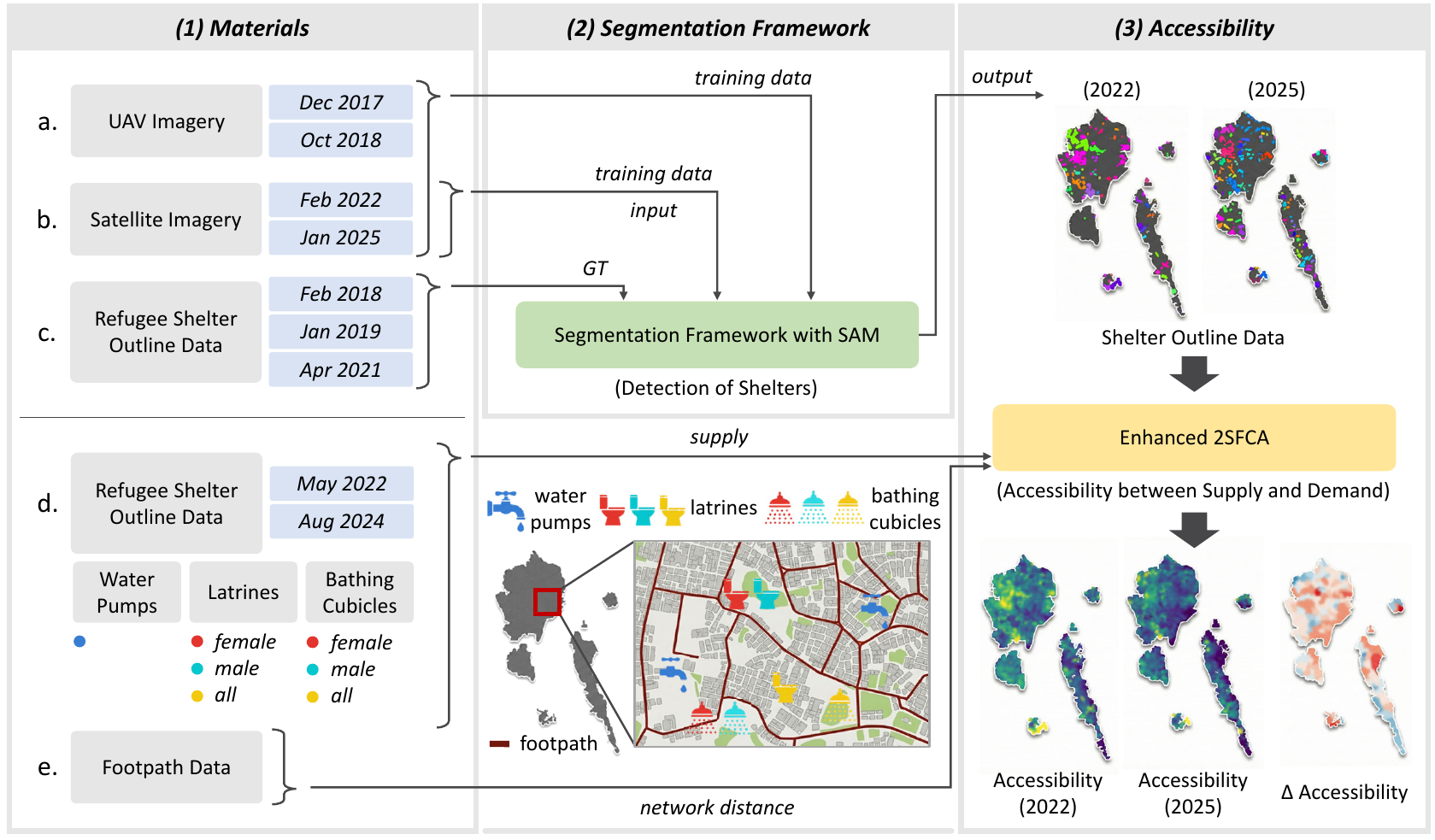}
\caption{\textbf{Overview of the AI-driven Workflow for Refugee Shelter Mapping and WASH Accessibility Analysis.} The workflow consists of three stages: \textbf{(1) Materials}: Integration of multi-temporal UAV (\textbf{a}) and Satellite (\textbf{b}) imagery alongside refugee shelter outline data (GT) (\textbf{c}) for model training. \textbf{(2) Segmentation Framework}: Application of the proposed semi-supervised segmentation model based on SAM to detect shelter outlines for \textit{2022} and \textit{2025} from satellite imagery, serving as a granular spatial basis for estimating service demand. \textbf{(3) Accessibility}: Assessment of the spatial balance between demand (estimated from detected shelters) and supply (WASH facilities: water pumps, latrines, and bathing cubicles, including gender-disaggregated attributes) (\textbf{d}) using the enhanced 2SFCA method with network distances derived from footpath data (\textbf{e}).}
\label{fig:overview}
\end{figure*}

Remote sensing imagery offers a scalable alternative to these limitations, enabling extensive and continuous monitoring. Recently, computer vision segmentation algorithms have emerged as effective tools for analyzing such imagery, demonstrating robust performance in identifying structural elements like refugee shelters \cite{gao2024leveraging,Ahn25AAAI}. However, unlike the organized layouts of typical UN-managed camps \cite{Weigand25_IDPstructure}, the spontaneous growth of Rohingya camps on space-constrained terrain results in dense and irregular shelter patterns. Under these conditions, conventional segmentation models often merge adjacent shelters or miss small and occluded ones, significantly degrading overall accuracy.

This study introduces a semi-supervised segmentation framework based on the Segment Anything Model (SAM) \cite{kirillov2023segment} to map refugee shelters from remote sensing imagery and assesses WASH facility accessibility in Cox’s Bazar (\cref{fig:overview}). SAM is a promptable foundation model that produces masks for individual shelters from cues (e.g., points, boxes, or masks), even in visually complex scenes. We implement a student–teacher paradigm with pseudo-label refinement, converting teacher predictions into spatial prompts for SAM and integrating both outputs to obtain reliable pseudo-labels from unlabeled imagery. The framework draws on manually digitized shelter outlines from international humanitarian mapping actors (i.e., UNOSAT and the REACH) \cite{outlinedata} as ground-truth (GT) labels, paired with corresponding Unmanned Aerial Vehicle (UAV) and Very High-Resolution (VHR) satellite imagery, alongside additional unlabeled imagery covering \textit{2017} to \textit{2025}. Using these detailed maps, we quantify changes in living space per person and population density over time. We then integrate gridded population estimates with locations of WASH facilities (i.e., water pumps, latrines, and bathing cubicles) and apply an enhanced Two-Step Floating Catchment Area (2SFCA) method \cite{luo2009enhanced,bryant2019examination} to evaluate spatial mismatches between demand and supply for \textit{2022} and \textit{2025}.

Our findings present that, while shelter areas largely stabilized after \textit{2020}, population continued to increase through \textit{2025}, reducing living space and intensifying overcrowding. Meanwhile, accessibility to WASH facilities has declined as demand fails to match facility supply, with users per facility exceeding global refugee standards. Notably, this shortfall is not evenly distributed across genders, as survey-based scenarios accounting for women's safety concerns with shared facilities reveal pronounced disparities, with female accessibility being lower than that for males. Ultimately, we translate granular geospatial information into actionable insights for humanitarian aid, serving as an evidence-based foundation to guide targeted interventions and promote equity in protracted displacement settings. 


\section*{Results}
\label{sec:results}
\subsection*{Segmentation Performance Evaluation}
\label{sec:segmentation_performance_evaluation}
We evaluate the proposed semi-supervised segmentation framework on a multi-temporal dataset, comprising UAV and satellite imagery with sparsely available GT labels (Materials and Methods). Our framework is compared with three baseline categories: a supervised Fully Convolutional Network (FCN) \cite{long2015fcn}, previously applied to refugee shelter detection \cite{Ghorbanzadeh22,wernicke2023deep}; (ii) standard semantic segmentation methods, including supervised U-Net \cite{ronneberger2015unet} and UNetFormer \cite{wang2022unetformer}, and semi-supervised UniMatch-V2 \cite{yang2025unimatch}; and (iii) test-time adaptation methods, including TENT \cite{wang2021tent} and SHOT \cite{liang2020shot}, for domain shift robustness (\hyperref[sec:Segmentation_Framework_Implementation]{\textit{SI Appendix}, Text~\ref{sec:Segmentation_Framework_Implementation}}).

For quantitative assessment, we report conventional segmentation metrics—Intersection over Union (IoU), precision, recall, and F1-score (\Cref{tab: main_result}). The supervised baselines (FCN, U-Net, and UNetFormer) consistently exhibit lower precision but higher recall, suggesting that they tend to include non-shelter surroundings in visually complex scenes. Conversely, the semi-supervised baselines (UniMatch-V2, TENT, and SHOT) generally achieve higher precision, with F1-scores that are typically superior or comparable to those of the supervised baselines. This improvement in precision derives from how these methods exploit unlabeled data. Nevertheless, these baselines rely on pseudo-labels for supervision signals without an explicit denoising mechanism, leading to error accumulation during training. Our method refines pseudo-labels using SAM’s instance-level priors (\hyperref[sec:Effectiveness_Pseudo-label_refinement]{\textit{SI Appendix}, Text~\ref{sec:Effectiveness_Pseudo-label_refinement}}), yielding the best overall performance. It surpasses the refugee-related method (FCN) by 4.9$\%$ and even the second-best method (TENT) by 3.0$\%$ in F1-score.

\begin{figure*}[t!]
\centering
\includegraphics[width=14.5cm]{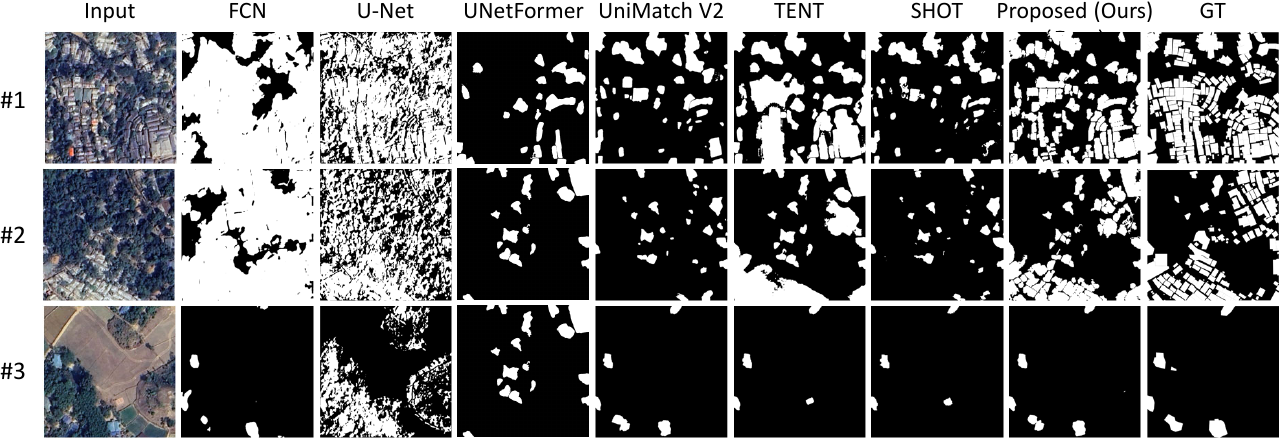}
\caption{\textbf{Qualitative Assessment for Refugee Camp Segmentation.} Representative results for high-density ($\#\textbf{1}$), mid-density ($\#\textbf{2}$), and low-density ($\#\textbf{3}$) camps, comparing our method with baselines. Our model yields masks closest to GT labels, effectively managing various noise or fragmentation from dense and irregular shapes. Satellite imagery sourced from the World Imagery Wayback data archive (\textit{2025}).}
\label{fig:segmentation_result}
\end{figure*}

To complement the quantitative results, we perform qualitative assessment with baselines on unlabeled data collected in \textit{2025} (\cref{fig:segmentation_result}). Three refugee shelter-density scenarios such as high-density camp ($\#\textbf{1}$), mid-density camp ($\#\textbf{2}$), and low-density camp ($\#\textbf{3}$) are considered. In high- and mid-density cases ($\#\textbf{1}$ and $\#\textbf{2}$), most baselines produce inaccurate masks due to severe spatial congestion and disordered shelters. Specifically, the supervised baselines (FCN, U-Net, and UNetFormer) generate noisy and fragmented predictions, frequently misclassifying vegetation as shelters. While the semi-supervised baselines (UniMatch-V2, TENT, and SHOT) reduce false positives, they do not resolve fine-grained details, commonly blending nearby shelters or overlooking compact or partially hidden features. In contrast, our method accurately separates closely spaced shelters while preserving sharp boundaries, even in heavily cluttered scenes. In the low-density case ($\#\textbf{3}$), the baselines except U-Net remain reliable. However, our method still achieves the fewest false negatives and the sharpest boundaries. These results highlight that our approach offers clear advantages in challenging high- and mid-density environments of the Rohingya camps.

\begin{table}[t!]
\centering
\caption{\textbf{Segmentation performance on the refugee camp shelter class, evaluated using intersection over union (IoU), precision, recall, and F1-score across all baselines.} Best scores are highlighted in bold with underline (\hyperref[sec:Segmentation_Framework_Implementation]{\textit{SI Appendix}, Text~\ref{sec:Segmentation_Framework_Implementation}}).}
\label{tab: main_result}
\setlength{\tabcolsep}{4pt}
\renewcommand{\arraystretch}{1.2}
\begin{tabular}{lcccc}
\toprule
\textbf{Methods} & \textbf{IoU (\%)} & \textbf{Precision (\%)} & \textbf{Recall (\%)} & \textbf{F1-score (\%)} \\
\midrule
FCN~\cite{long2015fcn}      & 61.5 & 65.2 & \textbf{\underline{79.1}} & 71.5 \\
U-Net~\cite{ronneberger2015unet}   & 60.1 & 63.8 & 78.4 & 70.3 \\
UNetFormer~\cite{wang2022unetformer}      & 63.8 & 68.5 & 78.8 & 73.3 \\
UniMatch-V2~\cite{yang2025unimatch} & 65.0 & 71.0 & 73.2 & 72.1 \\
TENT~\cite{wang2021tent}   & 66.5 & 71.5 & 75.4 & 73.4 \\
SHOT~\cite{liang2020shot}   & 65.8 & 72.3 & 73.5 & 72.9 \\
\textbf{Proposed (Ours)}      & \textbf{\underline{69.2}} & \textbf{\underline{75.8}} & 77.0 & \textbf{\underline{76.4}} \\
\bottomrule
\end{tabular}
\end{table}

\subsection*{Analysis of Rohingya Camp Growth and Living Space}
\label{sec:analysis_growth_living_space}
\begin{figure*}[t!]
\centering
\includegraphics[width=13cm]{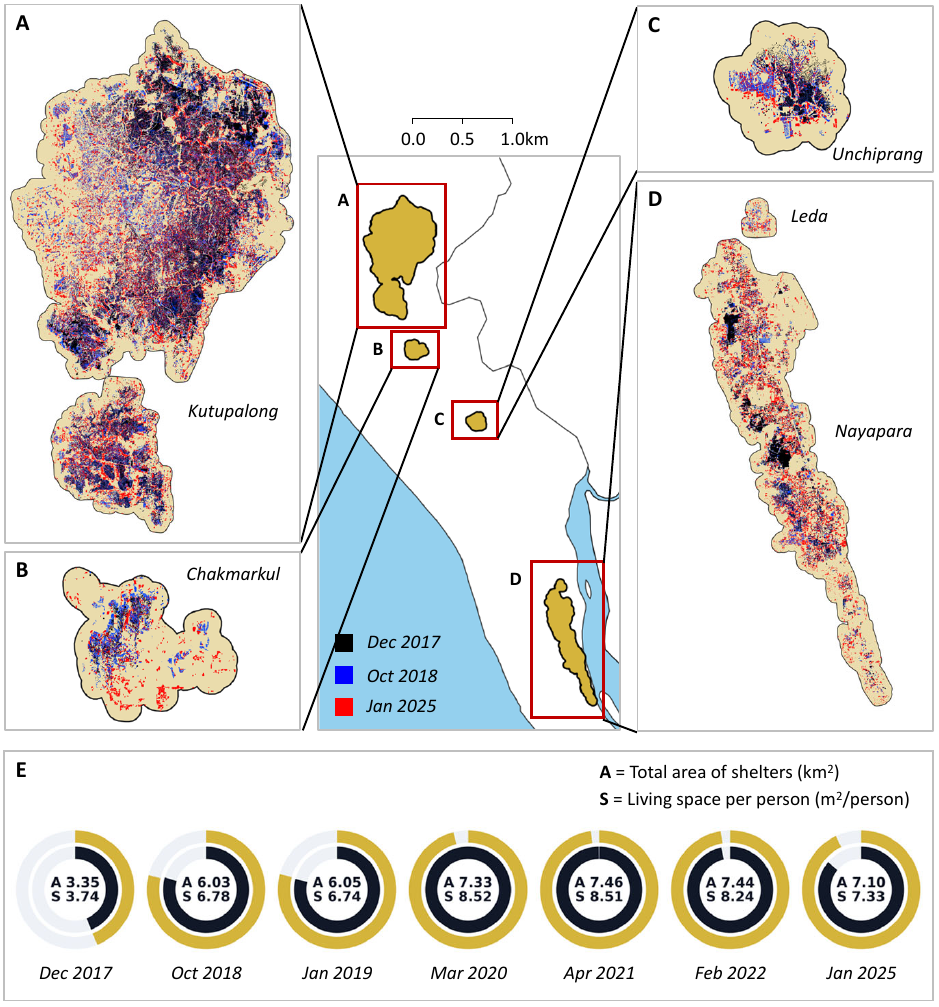}
\caption{\textbf{Analysis of Rohingya Camp Growth and Living Space in Cox’s Bazar.} (\textbf{A–D}) Detected refugee shelters in \textit{Dec 2017} (black), \textit{Oct 2018} (blue), and \textit{Jan 2025} (red). (\textbf{E}) Concentric rings across seven time points, estimated by our model (\textit{2017}, \textit{2018}, \textit{2019}, \textit{2022}, and \textit{2025}) and GT data (\textit{2020} and \textit{2021}) \cite{outlinedata}. The yellow outer and black inner rings show the total shelter area and per-capita living space indices (\textit{2017}=100), respectively.}
\label{fig:analysis_growth_living_space}
\end{figure*}

We examine evolving living conditions across 33 Rohingya refugee camps in Cox’s Bazar District using our segmentation framework to map spatial and demographic changes (\cref{fig:analysis_growth_living_space}). Between \textit{2017} and \textit{2025}, a consistent outward shift in shelter detections toward peripheral areas is observed (\hyperref[fig:analysis_growth_living_space]{Fig.~\ref*{fig:analysis_growth_living_space}A--D}). To quantify these longitudinal trends, we track the relationship between shelter area expansion and population density through a series of temporal indices (\hyperref[fig:analysis_growth_living_space]{Fig.~\ref*{fig:analysis_growth_living_space}E}).

The plots in \hyperref[fig:analysis_growth_living_space]{Fig.~\ref*{fig:analysis_growth_living_space}E} show surge in camp growth, with total shelter area nearly doubling between \textit{2017} and \textit{2018}. This rapid expansion coincides with the peak of the \textit{2017} Rohingya Crisis, where large-scale displacement from Myanmar overwhelmed existing infrastructure. During this initial phase, living space was restricted to a mere 3.74$m^2$ per person, underscoring severe constraints on basic needs such as sleeping, sanitation, and hygiene. Following a period of relative stabilization after \textit{2020}, the total shelter area recorded a modest decline. This trend likely reflects international efforts to formalize camps and enhance infrastructure, alongside localized losses due to fire events \cite{hassan2022mapping}. Despite this stabilization, refugee population growth persists, reducing per capita living space through \textit{2025} (8.52 $m^2$ in \textit{2020}; 7.33 $m^2$ in \textit{2025}). Although the \textit{2025} value remains above UNHCR emergency standards for tropical climates (3.5 $m^2$ per person) \cite{UNHCR25_IDP}, it falls short of OECD country norms (8–10 $m^2$ per person). This mismatch between static space and rising population heightens density, with implications for reduced service access from overcrowding.

\subsection*{Analysis of Accessibility to WASH Facilities}
\label{sec:analysis_accessibility}
\begin{figure*}[t!]
\centering
\includegraphics[width=14.5cm]{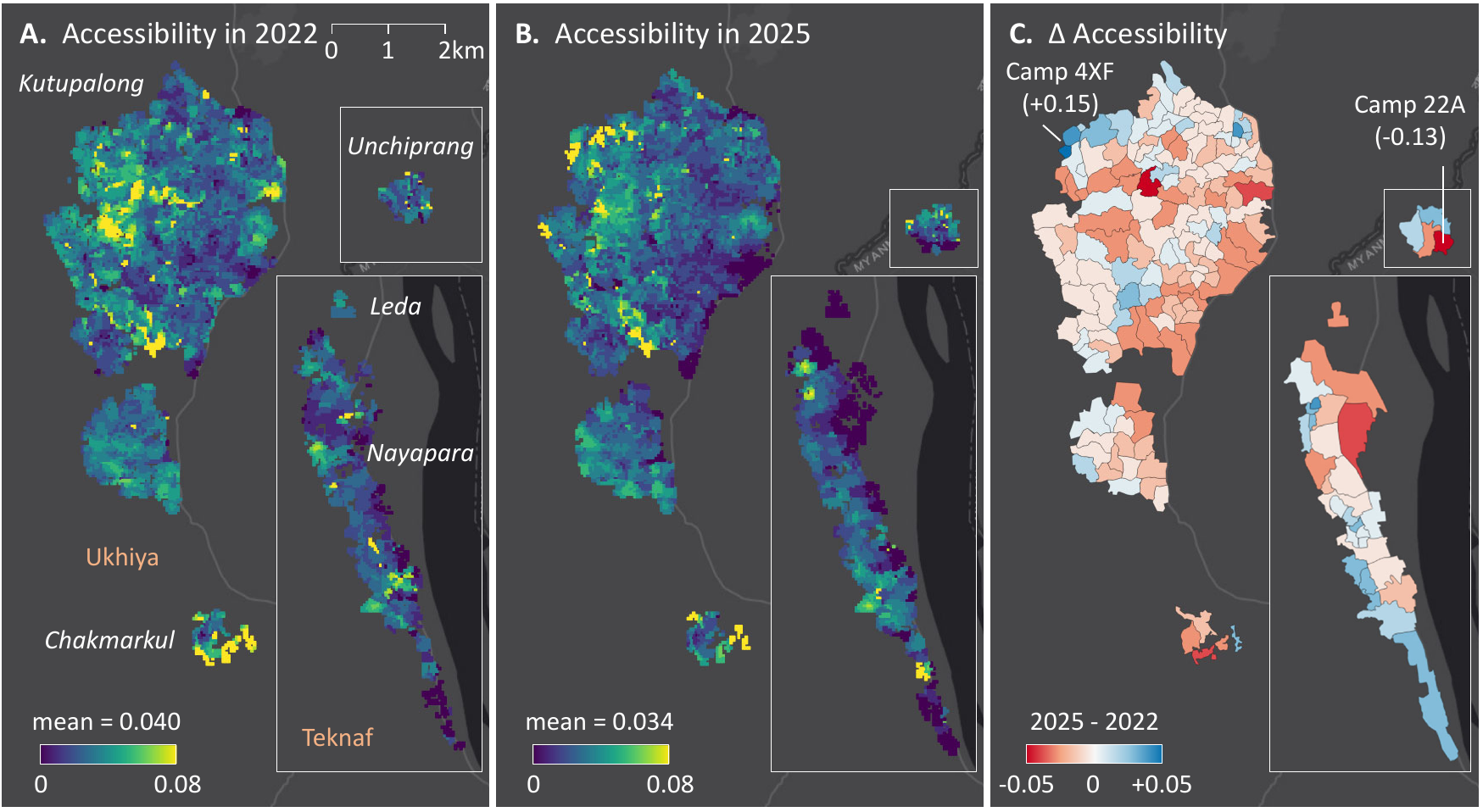}
\caption{\textbf{Accessibility to WASH Facilities based on Network Distances in \textit{2022} (A) and \textit{2025} (B), and Their Change (C).} Accessibility scores represent the mean across water pumps, latrines, and bathing cubicles. Changes are summarized at the administrative block level within camps, with blue indicating increases and red indicating decreases.}
\label{fig:accessibility_network_distance}
\end{figure*}

We compute accessibility from demand to supply using the 2SFCA method on a 50-meter grid with network distances (\cref{fig:accessibility_network_distance}). The demand is estimated from detected shelters weighted by grid-level population, and the supply is measured as the number of WASH facilities (\hyperref[sec:ACC_Implementation_details]{\textit{SI Appendix}, Text~\ref{sec:ACC_Implementation_details}}). An accessibility score is calculated separately for each type of WASH facility (i.e., water pumps, latrines, and bathing cubicles) and these types are then averaged to a single general accessibility value for \textit{2022} and \textit{2025}. For reference, Euclidean distance-based accessibility yields similar trends with the network-based one but less local detail (\hyperref[supply:s1_accessibility_euclidean_distance]{\textit{SI Appendix}, Fig.~\ref*{supply:s1_accessibility_euclidean_distance}})
.

At the camp level, average WASH accessibility decreases by $\sim$0.006, from 0.040 in \textit{2022} (\hyperref[fig:accessibility_network_distance]{Fig.~\ref*{fig:accessibility_network_distance}A}) to 0.034 in \textit{2025} (\hyperref[fig:accessibility_network_distance]{Fig.~\ref*{fig:accessibility_network_distance}B}). This equates to 1,000 people accessing 40 facilities in \textit{2022} versus 34 in \textit{2025}—or, inversely, each facility serving 25 people in \textit{2022} versus 29.4 in \textit{2025}. Over three years, facilities thus serve $\sim$4 more people (up $\sim$18$\%$), signaling marked deterioration in WASH conditions in Cox’s Bazar. By facility type (\hyperref[supply:s2_accessibility_network_distance]{\textit{SI Appendix}, Fig.~\ref*{supply:s2_accessibility_network_distance}}), a water pump serves 67.1 people in \textit{2025} (up from 47.8 in \textit{2022}); a latrine serves 17.2 people (up from 15.0); and a bathing cubicle serves 32.9 people (up from 30.7). Among facilities, the \textit{2025} latrine load exceeds the \textit{2024} global average of 24 refugees per latrine across 104 settlements in 17 countries \cite{UNHCR24_WASH}.

At the administrative block level within camps, accessibility changes are spatially heterogeneous, visualized with red for decreases and blue for increases (\hyperref[fig:accessibility_network_distance]{Fig.~\ref*{fig:accessibility_network_distance}C}). The largest drops occur in eastern Kutupalong and southern Unchiprang (Camp 22A), with an average decrease of 0.12 ($\sim$13 more people per facility). In contrast, the largest rises occur in northern Kutupalong (Block F of Camp 4X), with an average increase of 0.15 ($\sim$14 fewer people per one facility). Notably, areas with the greatest drops did not exhibit the highest population growth; rather, the primary driver was the minimal addition of facilities (\hyperref[supply:s3_population_density]{\textit{SI Appendix}, Fig.~\ref*{supply:s3_population_density}}).

To validate our accessibility scores, we compare block-averaged results (\textit{Jan 2025}) with camp-level survey data on people per functional latrine from the Rohingya Refugee Response (RRR) WASH Sector Gap Analysis (\textit{sep 2024}) as GT for local access to sanitation facilities across 33 camps. The comparison reveals a strong Spearman’s correlation ($\rho$ = 0.709), demonstrating the empirical validity of our geospatial framework in mirroring the nuanced distribution of facility loads reported by on-the-ground evaluations (\hyperref[supply:s4_spearman]{\textit{SI Appendix}, Fig.~\ref*{supply:s4_spearman}}).

\begin{table*}[t!]
\centering
\caption{Summary statistics for three types of WASH facilities in \textit{2022} and \textit{2024}.}
\label{table:summary}
\setlength{\tabcolsep}{6pt}
\renewcommand{\arraystretch}{1.2}
\footnotesize
\begin{tabular}{l|ccc|ccc}
\toprule
                          & \multicolumn{3}{c}{\textbf{\textit{May 2022}}} & \multicolumn{3}{|c}{\textbf{\textit{Aug 2024}}} \\
                          & \textbf{Water pump} & \textbf{Latrine} & \textbf{Bath cubicle} 
                          & \textbf{Water pump} & \textbf{Latrine} & \textbf{Bath cubicle} \\
\midrule
Total locations   & 17,639 & 34,343 & 19,623 & 15,620 & 37,057 & 20,686 \\
Total facilities  & 17,661 & 52,635 & 25,974 & 15,620 & 52,495 & 25,805 \\
Facilities per location   & 1.01   & 1.53   & 1.32   & 1.00   & 1.42   & 1.25   \\
Facilities for female         & --     & 6,284  & 4,504  & --     & --     & --     \\
Facilities for male           & --     & 5,396  & 395    & --     & --     & --     \\
Facilities for all genders    & --     & 40,955 & 21,075 & --     & --     & --     \\
Facilities per 100 people     & 1.96   & 5.84   & 2.89   & 1.63   & 5.48   & 2.69   \\
\bottomrule
\end{tabular}
\begin{minipage}{\textwidth}
\footnotesize\raggedright
Note: ``Location'' refers to a geographic point with one or more WASH facilities. Each ``Facility'' is defined as a single unit of infrastructure (e.g., a tube well, a latrine, or a bathing cubicle). The facility count per location reflects infrastructure scale.
\end{minipage}
\end{table*}

\subsection*{Gender Disparities in WASH Accessibility}
\label{sec:gender_disparities}
\begin{figure*}[t!]
\centering
\includegraphics[width=14.5cm]{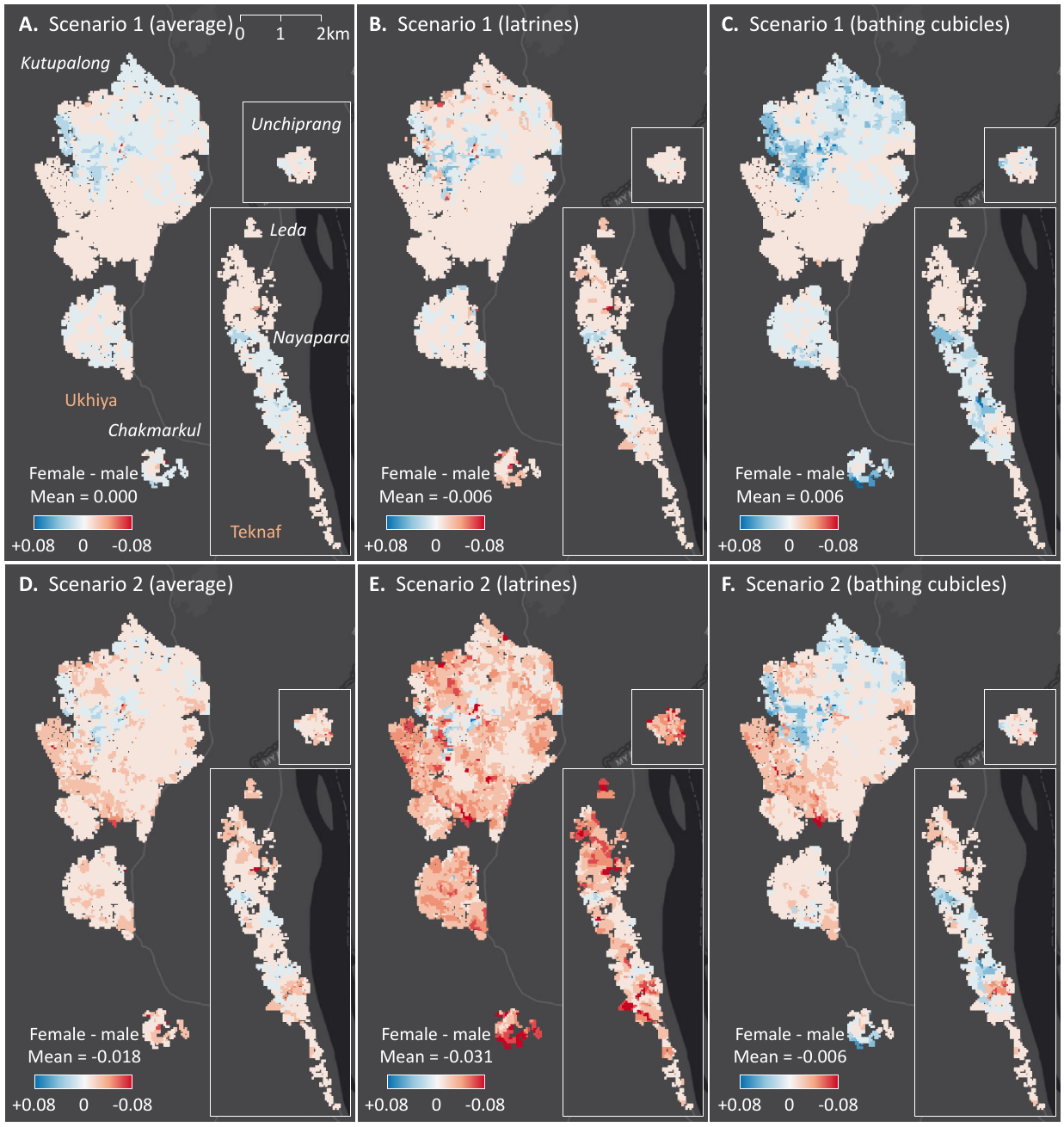}
\caption{\textbf{Gender-disaggregated WASH Accessibility Analysis.} Average in three facilities (\textbf{A}, \textbf{D}), latrines (\textbf{B}, \textbf{E}), and bathing cubicles (\textbf{C}, \textbf{F}). Blue and red indicate relatively higher accessibility for females and males, respectively. Scenario 1 (\textbf{A–C}) assumes that females are not reluctant to use all-gender facilities. Scenario 2 (\textbf{D–F}) assumes that 25$\%$ of females are reluctant to use all-gender facilities, based on thresholds derived from survey data \cite{Rohingya_WASH_dashboard}.}
\label{fig:gender_disaggregated}
\end{figure*}

We report gender-disaggregated WASH accessibility under real-world scenarios (\cref{fig:gender_disaggregated}). These differences arise from uneven provision of female- and male-designated facilities and demographic distributions. In \cref{fig:gender_disaggregated}, blue and red denote higher relative accessibility for females (i.e., women and girls) and males (i.e., men and boys), respectively. To capture potential constraints on women's use of shared facilities, we define two scenarios for female FDMNs' utilization of all-gender WASH facilities. Scenario 1 (\hyperref[fig:gender_disaggregated]{Fig.~\ref*{fig:gender_disaggregated}A–C}) considers unrestricted female access to all-gender facilities. In contrast, Scenario 2 (\hyperref[fig:gender_disaggregated]{Fig.~\ref*{fig:gender_disaggregated}D–F}) applies a 25$\%$ penalty to the effective capacity of all-gender facilities for females, informed by \textit{2022} survey data indicating that 25$\%$ of females in Cox’s Bazar feel unsafe using communal latrines at night due to risks of gender-based violence \cite{Rohingya_WASH_dashboard}.

In Scenario 1, the aggregated accessibility is similar across genders (0.0661 for females vs. 0.0659 for males; \hyperref[fig:gender_disaggregated]{Fig.~\ref*{fig:gender_disaggregated}A}). Latrine accessibility (\hyperref[fig:gender_disaggregated]{Fig.~\ref*{fig:gender_disaggregated}B}) is slightly lower for females than for males (0.115 vs. 0.121), despite a larger number of female-designated latrines (6,284 vs. 5,396; \Cref{table:summary}), reflecting the larger female population share (51.5$\%$ in the refugee camp population) \cite{UNHCR25site}. Bathing cubicle accessibility (\hyperref[fig:gender_disaggregated]{Fig.~\ref*{fig:gender_disaggregated}C}) is higher for women than males (0.062 vs. 0.056), aligned with the greater number of female-designated cubicles (4,505 vs. 39; \Cref{table:summary}). However, most cubicles remain unsegregated (21,075; \Cref{table:summary}), suggesting that actual female access may fall short of Scenario 1’s estimates.

In Scenario 2, incorporating constraints on female use of all-gender facilities, a pronounced gender gap emerges (0.048 for females vs. 0.066 for males; \hyperref[fig:gender_disaggregated]{Fig.~\ref*{fig:gender_disaggregated}D}). This corresponds to higher facility loads for females on gender-segregated facilities (20.8 females per female-designated facility vs. 15.2 males per male-designated facility). By facility type, loads remain higher for females than males (latrines: 11.1 vs. 8.3; \hyperref[fig:gender_disaggregated]{Fig.~\ref*{fig:gender_disaggregated}E}; bathing cubicles: 20.3 vs. 18.0; \hyperref[fig:gender_disaggregated]{Fig.~\ref*{fig:gender_disaggregated}F}), underscoring substantial gender disparities in WASH service conditions in Cox’s Bazar. Spatially, relatively higher female accessibility is concentrated in central and northern Kutupalong (blue in \hyperref[fig:gender_disaggregated]{Fig.~\ref*{fig:gender_disaggregated}F}), whereas female accessibility declines toward peripheral areas (red in \hyperref[fig:gender_disaggregated]{Fig.~\ref*{fig:gender_disaggregated}E}).

\section*{Discussion}
\label{sec:discussion}
The observed evolution from outward expansion to internal saturation signifies a pivotal transition in the settlement’s spatial-temporal development. While the total area occupied by shelters has stabilized, the persistent reduction in per capita living space underscores a progression into a protracted phase where spatial exhaustion without further spatial expansion becomes the primary driver of public health risk. This saturation point puts a strain on the settlement’s physical framework, complicating the maintenance of basic infrastructure and further entrenching the vulnerabilities of the residing population.

The reliability of this spatial evidence is rooted in the framework’s ability to navigate the complex visual environment of spontaneous settlements. By integrating vision foundation model priors to refine supervision signals, the system effectively differentiates diverse shelters from environmental noise, preventing shelter merging and omission errors. This approach provides a precise demand-side census where manual annotations are cost-prohibitive. Such granular data is a prerequisite for moving beyond camp-wide averages toward a more rigorous, spatially-detailed understanding of infrastructure pressure.

The integration of this AI-derived demand mapping with external facility data further reveals a supply-demand mismatch in essential service provision. The decline in WASH accessibility—dropping from 40 to 34 facilities per 1,000 people between \textit{2022} and \textit{2025}—quantifies a widening discrepancy between humanitarian requirements and the available infrastructure. This deterioration reflects a chronic resource shortfall exacerbated by declining international aid, such as the defunding of major programs like USAID \cite{cavalcanti2025evaluating,UNICEF25_WASH}. Notably, the functional divide exposed through our safety penalty scenario suggests that proximity alone is an insufficient measure of accessibility. The approximately 27$\%$ lower accessibility for females highlights that uniform service planning inherently compromises protection, potentially increasing the risk of gender-based violence (GBV) in shared or overcrowded facilities \cite{GBV2025Q1}. Ensuring humanitarian dignity thus requires a systemic reconfiguration of WASH environment that prioritizes safety and inclusive design \cite{crawley2021gender,ezbakhe2019leaving}.

While this work provides a robust diagnostic tool, its scope remains subject to the limitations of exogenous supply data, which may not reflect the real-time operational status. Integrating ground-level behavioral data remains a necessary step for future improvements. Nevertheless, the patterns identified in Cox’s Bazar serve as a vital blueprint for longitudinal monitoring in other protracted displacement contexts, such as those in Kenya or Uganda \cite{calderon2022social}. By prioritizing potential applicability in its design \cite{wernicke2023deep}, this study offers a scalable foundation for evidence-based interventions in the world’s most vulnerable and unmapped environments.

\begin{figure*}[t!]
\centering
\includegraphics[width=13cm]{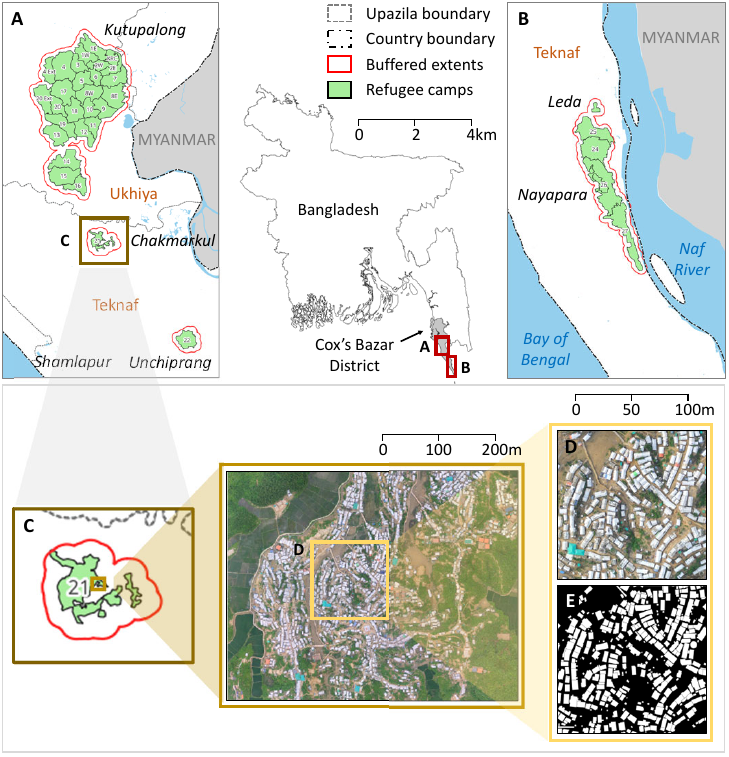}
\caption{\textbf{Map of the Study Area and Data Example.} (\textbf{A} and \textbf{B}) Geographic distribution of the 33 registered Rohingya refugee camps across Ukhiya and Teknaf \textit{Upazilas} of Cox’s Bazar District, Bangladesh. Green polygons denote the official camp boundaries (sourced from the Inter-Sector Coordination Group (ISCG), \textit{May 2024}), while red outlines indicate the 100-meter buffered extents defined as the strict analysis domain for the segmentation framework. (\textbf{C}) Representative UAV imagery of Camp 21 (Chakmarkul) illustrating typical shelter layouts. (\textbf{D} and \textbf{E}) Examples of model inputs: (\textbf{D}) a standardized imagery tile (Zoom Level 18) and (\textbf{E}) corresponding GT shelter footprints (sourced from the Humanitarian Data Exchange (HDX)).}
\label{fig:study_area}
\end{figure*}

\section*{Materials and Methods}
\label{sec:materials_methods}
\noindent\textbf{Study Area.} The study focuses on the Rohingya refugee camps located in the Ukhiya and Teknaf \textit{Upazilas} of Cox’s Bazar District, Bangladesh (\cref{fig:study_area}). After the \textit{2017} mass displacement from Myanmar, this region hosts one of the world's largest refugee concentrations, currently organized into 33 registered camps (green areas in \cref{fig:study_area}). The majority is clustered around Kutupalong in Ukhiya \cite{iDMC25_IDP}. To ensure longitudinal consistency and data reliability, we excluded Camp 23 (Shamlapur), following its relocation to Bhasan Char in \textit{Dec 2021}, as well as adjacent unregistered settlements where humanitarian interventions and facility records are not systematically collected and maintained. Accordingly, the analysis domain is strictly defined as the 100-meter buffered extents of the registered camps (red outlines in \hyperref[fig:study_area]{Fig.~\ref*{fig:study_area}A, B}).
\smallskip

\noindent\textbf{Refugee Shelter Outline Data and Imagery.} To map shelter dynamics, we integrated GT labels with multi-temporal remote sensing imagery.

\begin{itemize}
    \item \textbf{Refugee Shelter Outline Data (GT).} Geospatial building footprints were retrieved from the Humanitarian Data Exchange (HDX) \cite{outlinedata}. Derived from UAV imagery captured by the International Organization for Migration (IOM) and processed by UNOSAT and REACH, these data capture diverse structural features, including self-built bamboo, tarpaulin, and concrete shelters (\hyperref[fig:study_area]{Fig.~\ref*{fig:study_area}E}). Three temporal snapshots (\textit{Feb 2018}, \textit{Jan 2019}, \textit{Apr 2021}) were utilized as GT labels for model training and as additional data points for camp growth analysis. These labels were paired with corresponding imagery (\textit{Dec 2017}, \textit{Oct 2018}, \textit{Feb 2022}). Due to temporal offsets between image acquisition and labeling, a correction algorithm was applied to rectify minor spatial misalignments during the training process (\hyperref[sec:Alignment_via_STR_matching]{\textit{SI Appendix}, Text~\ref{sec:Alignment_via_STR_matching}}).
    \item \textbf{Remote Sensing Imagery.} A multi-source dataset comprising UAV and VHR satellite imagery was constructed. For the initial period, UAV imagery (native 10 cm/pixel resolution) was acquired from the IOM Needs and Population Monitoring initiative (\textit{Dec 2017}, \textit{Oct 2018}) \cite{UAVdata}. To extend temporal coverage, Maxar (GE01) satellite imagery was sourced via the ESRI World Imagery Wayback using the ESRI ArcGIS REST API (\textit{Feb 2022}), while Airbus satellite imagery was sourced via Google Earth Engine (\textit{Jan 2025}). To ensure consistency, all imagery was standardized to Zoom Level 18 ($\sim$0.60 m/pixel resolution) (\hyperref[fig:study_area]{Fig.~\ref*{fig:study_area}D}) and tiled into non-overlapping 256×256 pixel patches. This processing pipeline yielded a total of 7,406 UAV tiles and 7,406 satellite tiles, respectively.
\end{itemize}

\begin{figure*}[t!]
\centering
\includegraphics[width=14cm]{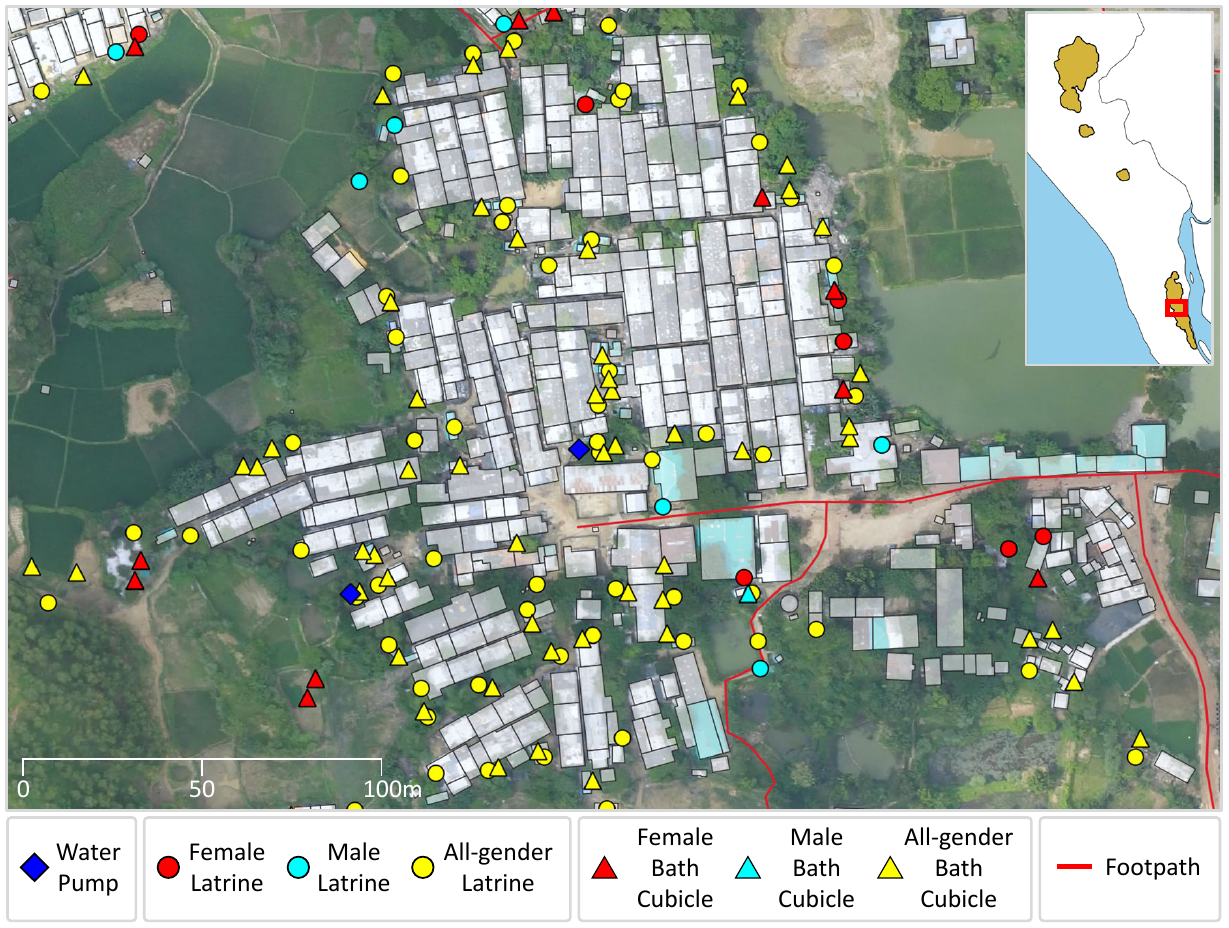}
\caption{\textbf{Neighborhood Map of Water Pumps (Quadrangles), Latrines (Circles), and Bathing Cubicles (Triangles) in \textit{2022} for Camp 26, Teknaf.} Latrines and bathing cubicles are classified by gender designation.}
\label{fig:neighborhood_map}
\end{figure*}

\noindent\textbf{WASH Facility Data.} Geospatial data on WASH infrastructure were obtained from the Rohingya Refugee Response (RRR) for \textit{May 2022} and \textit{Aug 2024} \cite{RRR25rohingya}. We restricted the analysis to facilities within registered camp boundaries, categorized into water pumps, latrines, and bathing cubicles (\cref{fig:neighborhood_map}). For temporal alignment with the remote sensing imagery, the \textit{May 2022} facility data were paired with \textit{Feb 2022} imagery, while the \textit{Aug 2024} data with \textit{Jan 2025} imagery. The dataset distinguishes between a “Location” (unique coordinates) and a “Facility” (individual physical unit), providing summary statistics of the infrastructure distribution (\Cref{table:summary}; \hyperref[sec:Rohingya_Data_Details]{\textit{SI Appendix}, Text~\ref{sec:Rohingya_Data_Details}}). Here, the \textit{2022} dataset specifically contains gender-disaggregated attributes, which serve as the basis for modeling gender-specific accessibility in conjunction with safety perception survey data \cite{Rohingya_WASH_dashboard}. 
\smallskip

\noindent\textbf{Footpath Data.} A pedestrian network was modeled using the Rohingya Refugee Response footpath data (\textit{sep 2025}) to calculate accessibility scores \cite{Rohingya_ROAD}. Shelter centroids and facility locations are snapped to the nearest network nodes or edges. Total distance is computed as the sum of the shortest on-network path (i.e., path along the network) and off-network offset (i.e., distance from the location to the network) distances.

\subsection*{Segmentation Framework for Refugee Shelter Detection}
\subsubsection*{Problem Formulation}
We define the shelter detection task within a semi-supervised framework using a small set of labeled data
$D_l=\{(x_i,y_i)\}_{i=1}^{N_l}$ and a larger set of unlabeled data $D_u=\{x_j\}_{j=1}^{N_u}$.
For the labeled data, each image $x_i\in\mathbb{R}^{H\times W\times 3}$ is paired with a corresponding binary mask
$y_i\in\{0,1\}^{H\times W}$, where pixels are annotated as either \textit{Background} (class 0) or \textit{Refugee Shelter} (class 1).
The primary objective is to train a segmentation model $f_{\theta}$ that estimates a binary mask $y$ from an image $x$.

\subsubsection*{Semi-Supervised Learning Framework}
\begin{figure*}[t!]
\centering
\includegraphics[width=14cm]{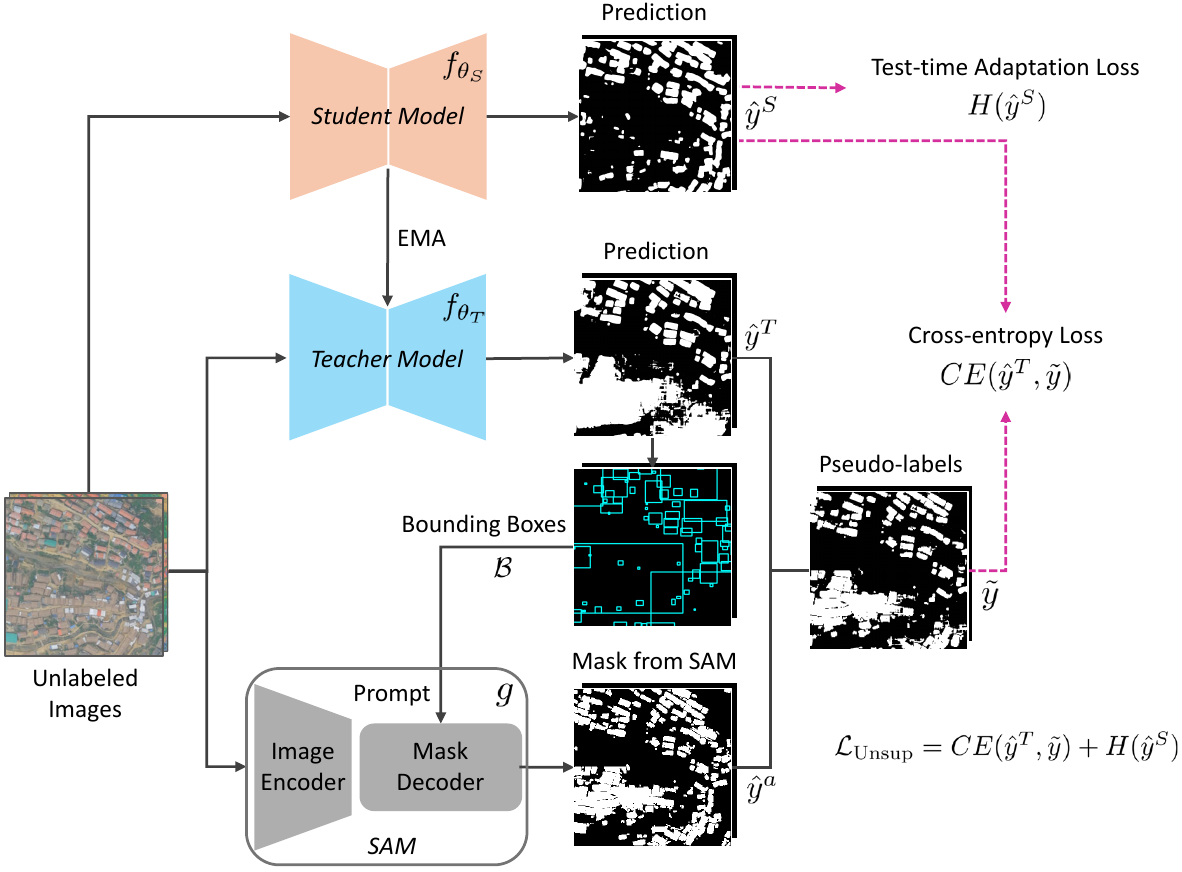}
\caption{\textbf{Overview of the Proposed Semi-supervised Segmentation Framework for Refugee Shelter Detection.} The teacher model $f_{\theta_T}$ is first trained on labeled data. It then generates coarse predictions $\hat{y}^{T}$ for unlabeled images, which are refined using a guided Segment Anything Model (SAM) $g$ with bounding-box prompts $B$ to produce accurate pseudo-labels $\tilde{y}$. The student model $f_{\theta_S}$ is optimized to minimize the loss function $L_{\mathrm{Unsup}}$, composed of a cross-entropy term $\mathrm{CE}$ between its predictions $\hat{y}^{S}$ and the refined pseudo-labels, with an entropy regularization term $H$ on $\hat{y}^{S}$ for domain adaptation. The teacher is updated via an exponential moving average of the student’s weights, progressively improving pseudo-label quality.}
\label{fig:segmentation_model}
\end{figure*}

Given the constraints of limited pixel-level annotations and domain shifts inherent in humanitarian remote sensing,
we propose a student-teacher paradigm enhanced by a promptable foundation model (\cref{fig:segmentation_model}).
This method operates on a pseudo-labeling approach where a teacher model provides initial guidance derived from unlabeled
data to a student model, enabling the student to learn more reliably by incorporating this guidance throughout the training
process. Each model takes an image as input $x$ and returns a probability map $p=f_{\theta}(x)$, which is converted into a binary
mask $\hat{y}$ via thresholding.

Initially, the teacher model $f_{\theta_T}$ is pre-trained on the labeled data $D_l$ by minimizing the supervised loss
$\mathcal{L}_{\mathrm{Sup}}$ via standard cross-entropy:
\begin{equation}
\mathcal{L}_{\mathrm{Sup}}=\frac{1}{N_l}\sum_{(x_i,y_i)\in D_l}\mathrm{CE}\!\left(f_{\theta_T}(x_i),y_i\right).
\end{equation}

Following this supervised phase, the teacher model produces coarse predictions $\hat{y}^{\,T}$ for the unlabeled data $D_u$.
To rectify initial errors, SAM $g$ refines these outputs into more accurate pseudo-labels $\tilde{y}$ that serve as reliable
supervision. Subsequently, the student model $f_{\theta_S}$ is optimized on $D_u$ to align its predictions with the refined
labels. To simultaneously bridge the domain shift arising from distribution gap between training and test data, we adopt an
entropy-based regularization strategy inspired by test-time adaptation. This dual objective constitutes the unsupervised loss
$\mathcal{L}_{\mathrm{Unsup}}$:
\begin{equation}
\mathcal{L}_{\mathrm{Unsup}}=\frac{1}{N_u}\sum_{x_j\in D_u}\left[\mathrm{CE}\!\left(f_{\theta_S}(x_j),\tilde{y}_j\right)
+H\!\left(f_{\theta_S}(x_j)\right)\right],
\end{equation}
where
\begin{equation}
H(p)=-\sum_{k} p_k \log p_k,
\end{equation}
with indexing $k$ classes, denotes the Shannon entropy \cite{baez2011characterization}.
To gradually improve pseudo-label quality, the teacher model is updated via an exponential moving average (EMA) of the student
model's parameters:
\begin{equation}
\theta_T \leftarrow \alpha\cdot \theta_T + (1-\alpha)\cdot \theta_S.
\end{equation}
Here, $\alpha$ is a decay rate typically set close to 1, and $\theta_T$ and $\theta_S$ are the teacher and student parameters,
respectively.

\subsubsection*{Pseudo-label Refinement with SAM}
To resolve the ambiguity in dense shelter arrangements where the teacher model often merges and misses instances, we leverage the
zero-shot generalization capabilities of SAM. While SAM offers high spatial precision, it requires effective prompts for accurate
localization. Accordingly, we extract bounding boxes $B$ from the prediction $\hat{y}^{\,T}_j$ of the teacher model
$f_{\theta_T}$ and use them as a spatial prompt for SAM $g$. This process generates a structure-aware mask $\hat{y}^{\,a}_j$
that respects object boundaries:
\begin{equation}
\hat{y}^{\,a}_j=g(x_j;B).
\end{equation}
The final pseudo-label $\tilde{y}_j$ is derived via the logical intersection of the teacher's prediction and the SAM-generated
mask:
\begin{equation}
\tilde{y}_j=\hat{y}^{\,T}_j \cap \hat{y}^{\,a}_j.
\end{equation}
This intersection functions as a consensus criterion, particularly in the early stages of training when coarse predictions may
yield oversized bounding boxes. By retaining only the regions where both models agree, we effectively filter out false positives
and provide rigorous supervision signals for the student model $f_{\theta_S}$.

\subsection*{Measuring Accessibility to WASH Facilities}
\subsubsection*{Problem Formulation}
To quantify spatial mismatches between service demand and facility supply, we apply an enhanced Two-Step Floating Catchment Area
(2SFCA) method incorporating a Gaussian distance-decay function. The method estimates accessibility by integrating population
needs and service availability within network-based catchments (\hyperref[sec:2SFCA_Background]{\textit{SI Appendix}, Text~\ref{sec:2SFCA_Background}}). Accessibility scores are computed on a
50-meter grid-level rather than at the individual-level using network distances to capture broader spatial trends.

Formally, let $u\in U$ denote the set of service demand locations (grid cell centroid) and $v\in V$ denote the set of facility
supply locations.
\begin{itemize}
  \item \textbf{Service Demand ($D_u$):} For each grid cell $u$, the service demand $D_u$ represents the estimated population size,
  derived by weighing the total shelter area by camp-specific demographic data \cite{UNHCR25site} (\hyperref[sec:ACC_Implementation_details]{\textit{SI Appendix}, Text~\ref{sec:ACC_Implementation_details}}).
  
  \item \textbf{Facility Supply ($S_v$):} For each facility $v$, the facility supply $S_v$ represents the functional service capacity,
  defined as the number of WASH facilities (i.e., water pumps, latrines, or bathing cubicles).
  \item \textbf{Distance ($d_{uv}$):} The network distance between locations $u$ and $v$ is denoted as $d_{uv}$ (\hyperref[sec:ACC_Implementation_details]{\textit{SI Appendix}, Text~\ref{sec:ACC_Implementation_details}}).
\end{itemize}

\subsubsection*{Spatial Interaction Modeling}
Spatial interaction is modeled using a Gaussian distance-decay function $W_{uv}$, which accounts for the friction of distance:
\begin{equation}
W_{uv}=
\begin{cases}
\exp\!\left(-\dfrac{d_{uv}^{2}}{\beta}\right), & \text{if } d_{uv}\le d_0,\\[6pt]
0, & \text{otherwise.}
\end{cases}
\end{equation}
where the threshold ($d_0$) is set to $1{,}609$ m (1 mile) as the maximum walking distance \cite{yang2012walking}. The decay coefficient $\beta$
is calibrated to induce a steep decline in weight at $402$ m (0.25 miles) \cite{bryant2019examination}. This function ensures that facilities closer
to a demand point contribute more to its accessibility score.

\subsubsection*{Accessibility Calculation via 2SFCA}
The accessibility calculation proceeds in two steps. First, we calculate the ratio of facility supply to service demand for each
facility $v$. This ratio ($R_v$) represents the service availability relative to the local demand:
\begin{equation}
R_v=\frac{S_v}{\sum_{k\in\{k\mid d_{kv}\le d_0\}} D_k\, W_{kv}},
\end{equation}
where the denominator aggregates the weighted service demand of all locations $k$ falling within the catchment of facility $v$.
A higher $R_v$ indicates lower competition for resources, while a lower $R_v$ indicates higher demand pressure relative to supply. Second, the
accessibility score ($A_u$) for each service demand location $u$ is obtained by summing the weighted ratios of all accessible
facilities:
\begin{equation}
A_u=\sum_{v\in\{v\mid d_{uv}\le d_0\}} R_v\, W_{uv},
\end{equation}
where $A_u$ quantifies the effective facility supply available to the population at location $u$. A higher $A_u$ indicates
better availability and proximity, while a lower $A_u$ indicates service scarcity and spatial remoteness.


\section*{Acknowledgments}
This work was supported by the Max Planck Institute for Security and Privacy, and by the National Research Foundation of Korea (NRF) through grants funded by the Korean government (MSIT) (No. RS-2022-00165347). Generative AI (GPT-5, Gemini) was used to edit and improve the clarity of the manuscript text; all content was reviewed and verified by the authors.

\bibliographystyle{unsrt}  
\bibliography{references}

\newpage
\begin{appendices}

\section*{Supporting Information Appendix (SI)}\label{sup:text}

\sitext{Implementation Details for Segmentation Framework} {sec:Segmentation_Framework_Implementation} \

\noindent\small\textbf{Training.} We configure UNetFormer \cite{wang2022unetformer} as the backbone for both teacher and student models. The labeled data consists of 2,418 UAV and 1,721 satellite images, while the unlabeled dataset consists of 4,988 UAV and 5,685 satellite images. For pre-training, we follow a 9:1 split on the labeled data. During training, images are randomly cropped to 768$\times$768 to improve diversity and robustness, while evaluation is performed on original size 1024$\times$1024 images. All pre-processing and data augmentation follow the official UNetFormer implementation. The teacher model is pre-trained for 40 epochs on the labeled data using AdamW with a learning rate of $1e-4$ and weight decay of 0.01. The student is then trained for an additional 40 epochs (80 in total) on unlabeled data with a reduced learning rate of $1e-5$, maintaining the same weight decay. During training, the teacher is updated via an exponential moving average (EMA) of the student’s parameters ($\alpha$=0.999). All experiments are conducted with a batch size of 32.
\smallskip

\noindent\small\textbf{Baselines.} We provide the architectural specifications and implementation details for the three baseline categories used in the comparative analysis. All baselines utilize the same dataset splits and preprocessing steps as the proposed framework to ensure fair comparability.
\begin{itemize}
    \item\small\textbf{Supervised Learning.} We benchmark against FCN-8s, a foundational architecture fusing multi-scale features, established as a standard baseline for refugee shelter mapping. We also employ U-Net, a standard encoder-decoder network utilizing symmetric skip connections to recover spatial details, and UNetFormer, a hybrid Transformer-CNN architecture capturing global context. Notably, UNetFormer serves as the backbone for both our framework and the test-time adapataion baselines to ensure comparable feature extraction.
    \item \small \textbf{Semi-Supervised Learning.} We use UniMatch-V2, a unified framework leveraging dual-stream perturbations. It exploits unlabeled data via weak-to-strong consistency regularization and FixMatch-style pseudo-labeling.
    \item \small \textbf{Test-Time Adaptation.} To address domain shifts, we implement TENT and SHOT. TENT updates affine batch-normalization parameters via entropy minimization on target data. SHOT optimizes the feature extractor via information maximization and pseudo-labeling while freezing the classifier.
\end{itemize}
\smallskip

\noindent\small\textbf{Evaluation Metrics.} To rigorously assess segmentation performance, we report Intersection over Union (IoU), Precision, Recall, and F1-score. All reported results are averaged over three random seeds for  statistical reliability.
\begin{itemize}
  \item \small \textbf{Intersection over Union (IoU).} Defined as the ratio of the intersection area to the union area
  $\left(\frac{|P \cap G|}{|P \cup G|}\right)$, it quantifies the spatial overlap between the predicted segmentation mask (P) and the ground-truth mask (G). It serves as the primary indicator of the geometric accuracy of detected shelter footprints.

  \item \small \textbf{Precision.} Calculated as the ratio of true positives to total predicted positives
  $\left(\frac{\mathrm{TP}}{\mathrm{TP}+\mathrm{FP}}\right)$, it assesses the reliability of detected shelters. It reflects the model's capability to differentiate shelters from environmental noise, thereby minimizing false positive errors.

  \item \small \textbf{Recall.} Calculated as the ratio of true positives to total actual positives
  $\left(\frac{\mathrm{TP}}{\mathrm{TP}+\mathrm{FN}}\right)$, it evaluates the completeness of shelter detection. It measures the capacity to identify all existing shelters within irregular camp environments, thereby minimizing false negative errors.

  \item \small \textbf{F1-score.} Calculated as the harmonic mean of Precision and Recall
  $\left(2 \times \frac{\mathrm{Precision}\times \mathrm{Recall}}{\mathrm{Precision}+\mathrm{Recall}}\right)$, it provides a balanced assessment of overall segmentation quality. It is particularly valuable for equally penalizing misclassification and missed detections.
\end{itemize}

\noindent \small \textbf{Hardware.} The experiment is performed on a single NVIDIA RTX 3090 GPU.

\sitext{Effectiveness of SAM-guided Pseudo-label Refinement} {sec:Effectiveness_Pseudo-label_refinement}

The impact of SAM-guided refinement is evaluated by tracking segmentation outputs throughout the training process. Representative results from high-density camp scenarios in both UAV and satellite imagery are provided (\hyperref[supply:s5_progressive_refinement]{\textit{SI Appendix}, Fig.~\ref*{supply:s5_progressive_refinement}}) indicate that the refinement mechanism progressively sharpens morphological boundaries and enhances the separation of clustered shelters. Notably, the framework recovers small or partially occluded structures that the initial teacher model failed to identify. Despite the inherent visual noise, the proposed method consistently preserves fine-grained structural details and reduces false positives. These results demonstrate that integrating instance-level priors from SAM effectively mitigates the limitations of conventional pseudo-labeling, providing the student model with rigorous supervision signals to navigate complex environments.

\sitext{Implementation details for Measuring Accessibility} {sec:ACC_Implementation_details} \

\noindent \small \textbf{Gridded Population.} Camp-level demographic data are allocated to 50-meter grid cells based on the detected shelters for quantifying service demand at a high geospatial granularity. This process begins by calculating the total area of shelters identified within each camp's administrative boundaries. Dividing the \textit{2022} and \textit{2025} population figures by these areas provides average population density (people per $m^2$ of shelter). To support gender-disaggregated accessibility, these calculations are performed separately for female, male, and total populations. The population for each grid cell is then determined by multiplying the density of the corresponding camp by the area of detected shelters within that cell. For grid cells intersecting multiple camp boundaries, population estimates are weighted according to the proportion of shelter area falling within each camp and then summed.

This approach is adopted because assessing accessibility based on the actual number of people offers a more intuitive view of how facilities are shared. By focusing on population rather than shelter area alone, the analysis more accurately reflects the real-world pressure on WASH services across the settlements.
\smallskip

\noindent\small\textbf{Network Distances.} An undirected network graph is constructed from footpath data, with each edge weighted by its physical length (m) to enable computation of minimum travel distances along the pedestrian network. Grid‑cell centroids (\textit{demand nodes}) and WASH facility locations (\textit{supply nodes}) are snapped to the nearest network vertex or edge to establish topological connectivity.

Total distance is defined as the sum of the shortest on-network path distance (the cumulative edge length along the graph between snapped locations) and the off-network offset distance (the Euclidean distance from the original point to its snapped location). In instances where the direct Euclidean distance between a demand and supply node is shorter than the combined offset distances—indicating immediate proximity not captured by the digitized network—the Euclidean value is utilized as a conservative proxy. Shortest-path matrices for all demand–supply pairs are calculated using the \textit{Parallel Hardware-Accelerated Shortest Path Trees (PHAST)} algorithm \cite{6012901}.
\smallskip

\noindent\small\textbf{Hardware and Software.} All geospatial analyses and accessibility modeling are implemented in R (v4.5.0) on an Ubuntu (24.04 LTS) environment equipped with 12 CPU cores and 64 GB of RAM. Core processing tasks use \textit{terra}, \textit{sf}, \textit{sfnetworks}, \textit{cppRouting}, and \textit{RcppParallel} libraries. All spatial data are projected to the WGS 84 / Pseudo-Mercator (EPSG:3857) coordinate system to maintain metric consistency for distance calculations.
\smallskip

\sitext{Alignment via SAM-based Structural Matching} {sec:Alignment_via_STR_matching}

Labeled data $\{x,y\}$ are subject to pre-processing alignment to rectify spatial discrepancies arising from sensor-induced geometric errors. This procedure utilizes masks generated by SAM $g$ as structural references to optimize label registration prior to training.

For each labeled image, a reference mask $y_{\text{ref}}$ is produced via $g$ to localize potential shelter boundaries. A discrete optimization is then performed over pixel translations $(\Delta x,\Delta y)$ and angular rotations $\phi$ to maximize the spatial agreement (F1-score) between the transformed ground-truth mask $y_{(\Delta x,\Delta y,\phi)}$ and $y_{\text{ref}}$.

\[
[\hat{\Delta x},\hat{\Delta y},\hat{\phi}]
=\arg\max_{(\Delta x,\Delta y,\phi)}
\;F1\!\left(y_{(\Delta x,\Delta y,\phi)},\,y_{\text{ref}}\right)
\]

This transformation is applied identically to both the image $x$ and its mask $y$. Specifically, empty regions vacated by the transformation are padded with zeros to maintain dimensional consistency. This approach facilitates spatial synchronization across diverse imagery by using SAM’s structural cues, even without extra training.

\sitext{Data Details: WASH Infrastructure and Demographic Dynamics} {sec:Rohingya_Data_Details} \

\noindent\small\textbf{Temporal Trends in Infrastructure Pressure.} Detailed summary statistics for the three types of WASH facilities are presented in \Cref{table:summary}. As described in the Materials and Methods, the data distinguishes between ``Locations'' and individual ``Facilities'' to reflect the actual scale of infrastructure. On average, each location consists of 1 water pump, 1.42–1.53 latrines, and 1.25–1.32 bathing cubicles.

The absolute number of all facility types decreased between \textit{2022} and \textit{2024}, a trend that quantifies the deterioration of essential service provision discussed in the main text. This decline is largely a consequence of a chronic resource shortfall linked to the defunding of major international programs \cite{cavalcanti2025evaluating,UNICEF25_WASH}. The resulting lack of maintenance and decommissioning of aging units has led to a significant supply-demand mismatch, especially as the refugee population grew from 899,867 to 958,614 during the same period. This reduction in per capita availability, despite international expansion efforts, underscores the settlement's progression into a protracted phase of spatial exhaustion.
\smallskip

\noindent\small\textbf{Gender-Disaggregated Provision and Safety Constraints.} While \textit{2022} data shows a higher number of female-designated facilities—reflecting an institutional commitment to inclusive design \cite{WASH2021tip}—a deeper functional divide exists within the WASH environment: 
\begin{itemize}
  \item \small \textbf{Infrastructure Imbalance.} The disproportionately low number of male-only bathing cubicles (395) compared to the vast majority of all-gender/unsegregated units (21,075) suggests that men primarily utilize shared facilities, whereas women’s access is more dependent on gender-specific zones.

  \item \small \textbf{Safety Penalty Modeling.} The safety penalty scenario (Scenario 2) is grounded in \textit{2022} survey data, which reveals that 25$\%$ of females feel unsafe using communal facilities at night \cite{Rohingya_WASH_dashboard}. This perceptual barrier effectively reduces the functional capacity of available infrastructure for females. By incorporating this penalty, our model accounts for the heightened risk of gender-based violence (GBV) in overcrowded settings, moving beyond camp-wide averages to provide a more equity-focused evaluation of humanitarian dignity.
\end{itemize}

\sitext{Background of 2SFCA} {sec:2SFCA_Background}

Over the past few decades, accessibility research has advanced through various metrics designed to quantify the spatial relationship between populations and essential infrastructure \cite{park2021review}. Early approaches relied on container-based metrics, which calculate facility-to-population ratios within administrative boundaries, or proximity-based metrics that consider distance to the nearest supply point. From the 1990s onward, gravity models gained prominence by integrating facility capacity and distance decay functions, offering a more nuanced representation of spatial interaction \cite{bryant2019examination}.

Building on this progression, the Two-Step Floating Catchment Area (2SFCA) method has been widely adopted for its balance of conceptual rigor and computational efficiency \cite{Chen19}. The primary strength of the 2SFCA framework lies in its dual-stage analysis, which systematically accounts for both supply and demand catchments. By evaluating these overlapping zones, the method effectively captures the local infrastructure pressure and the supply-demand mismatches inherent. The numerous extensions of 2SFCA have introduced diverse distance decay functions, adaptive catchment scales, and user-specific constraints \cite{luo2009enhanced}, transforming accessibility measurement into an increasingly sophisticated field of study.

\clearpage
\setcounter{figure}{0}
\renewcommand{\thefigure}{S\arabic{figure}}

\begin{figure*}[t!]
\centering
\includegraphics[width=14.5cm]{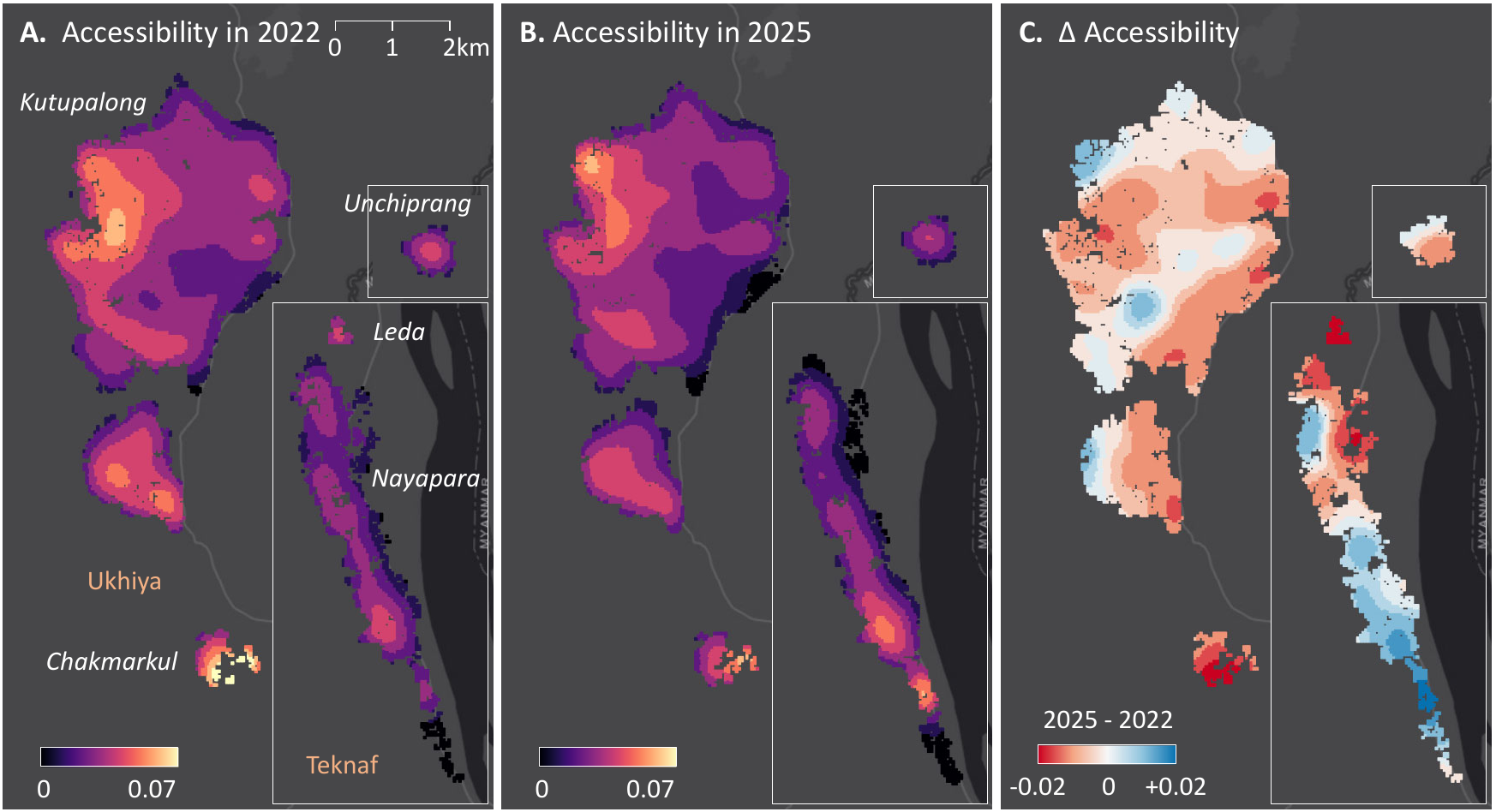}
\caption{\textbf{Accessibility to WASH Facilities based on Euclidean Distances in \textit{2022} (A) and \textit{2025} (B), and Their Change (C).} Accessibility scores represent the mean accessibility of water pumps, latrines, and bathing cubicles. Changes are summarized at the administrative block level within camps, with blue indicating increases and red indicating decreases.}
\label{supply:s1_accessibility_euclidean_distance}
\end{figure*}

\begin{figure*}[t!]
\centering
\includegraphics[width=14.5cm]{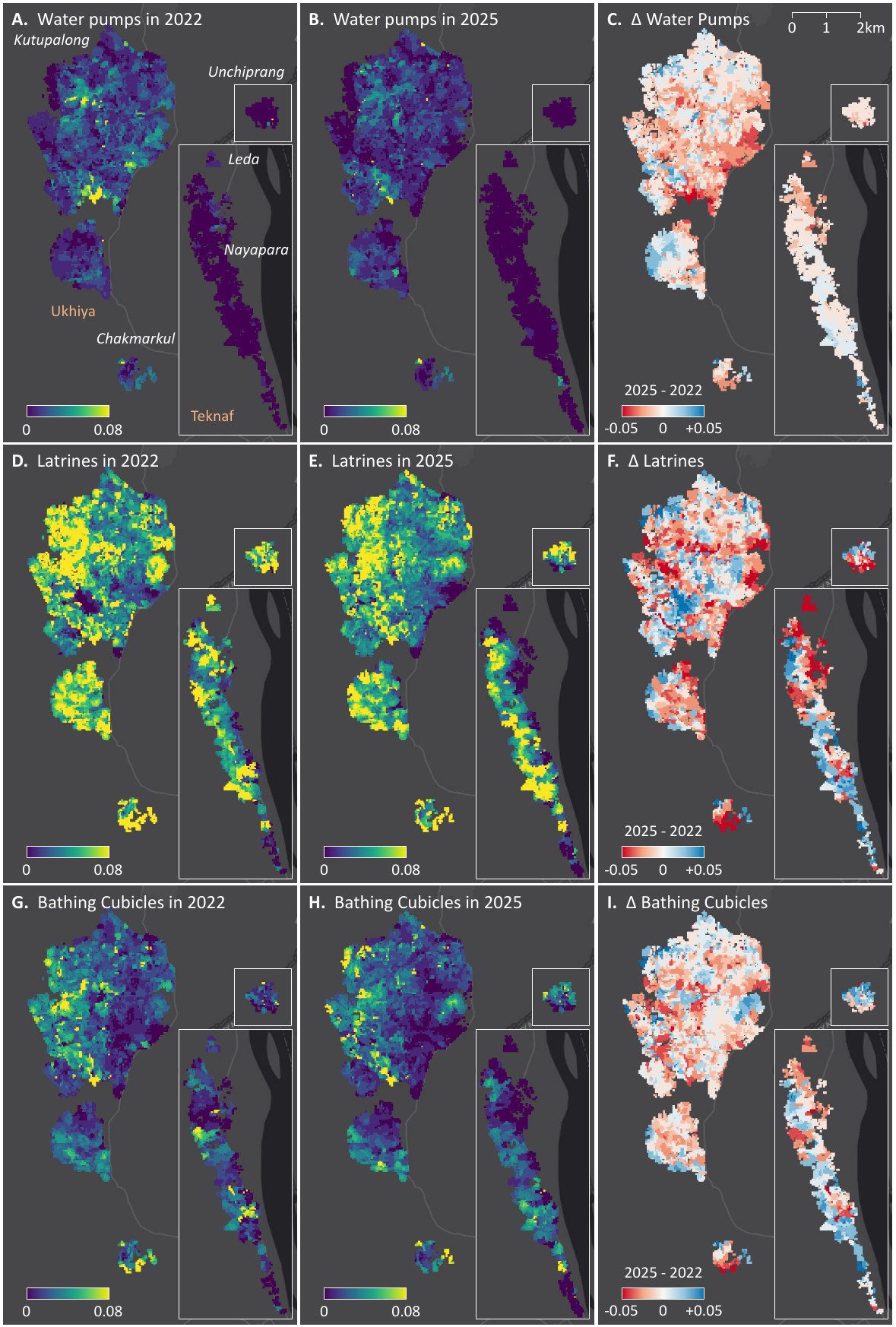}
\caption{\textbf{Accessibility based on Network Distances to Water Pumps (A–C), Latrines (D–F), and Bathing Cubicles (G–I) in \textit{202}, \textit{2025}, and Their Change.} Changes are summarized at the administrative block level within camps, with blue indicating increases and red indicating decreases.}
\label{supply:s2_accessibility_network_distance}
\end{figure*}

\begin{figure*}[t!]
\centering
\includegraphics[width=14.5cm]{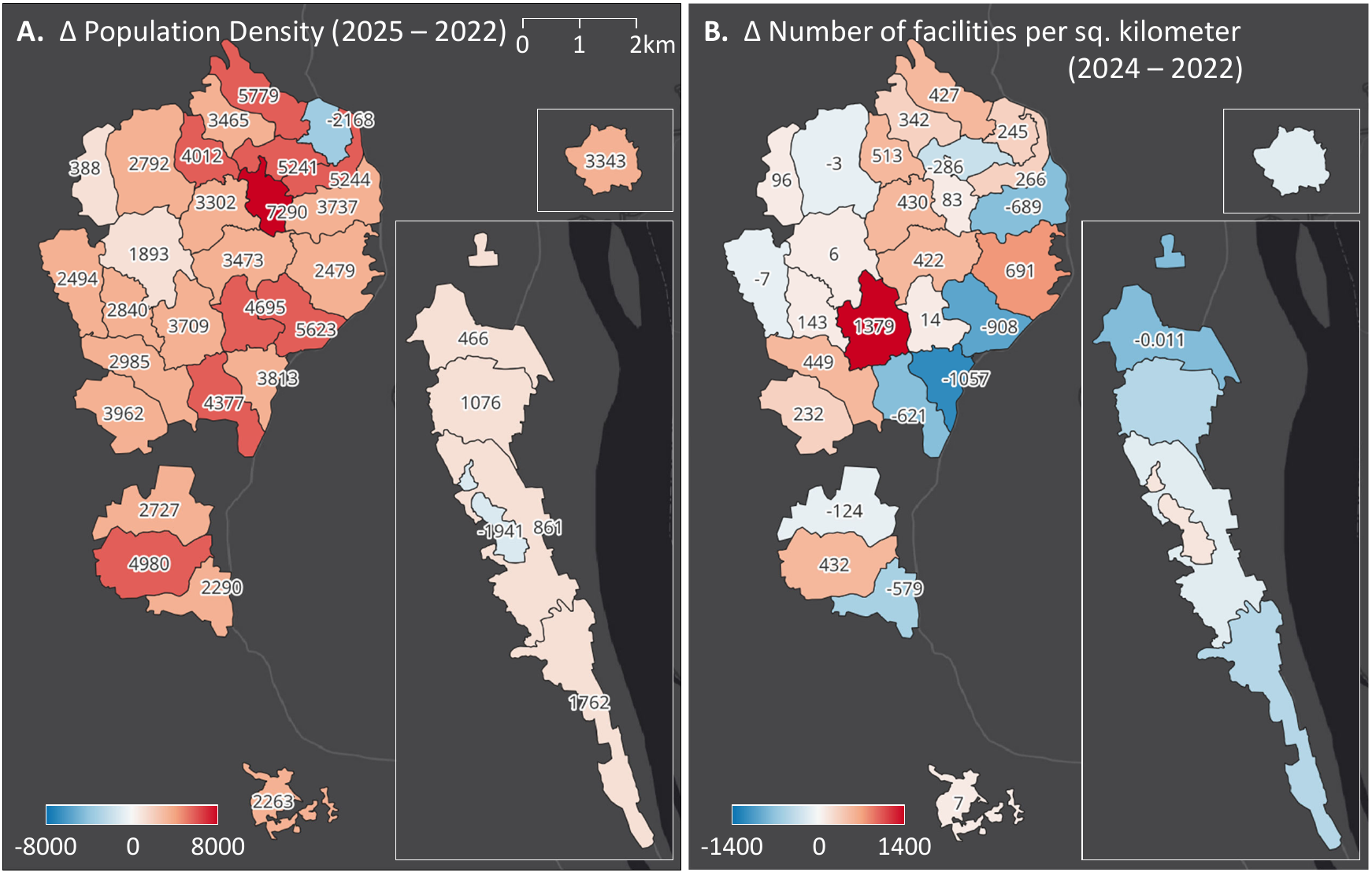}
\caption{\textbf{Changes in Population Density (A) and Facility Density (B) at the Camp Level.} Population density is compared between \textit{2022} and \textit{2025}, while facility density is compared between \textit{2022} and \textit{2024}, reflecting the respective data availability periods.}
\label{supply:s3_population_density}
\end{figure*}

\begin{figure*}[t!]
\centering
\includegraphics[width=9cm]{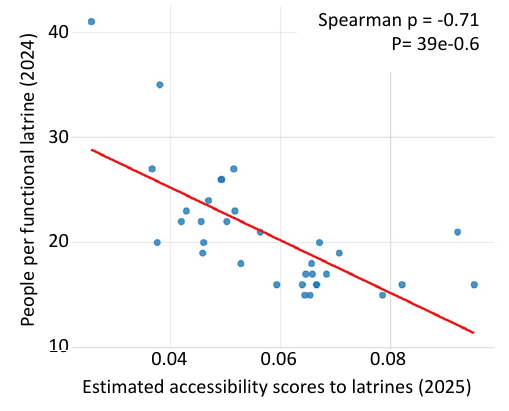}
\caption{\textbf{Validation of Accessibility Scores via Survey-Based Data.} Scatter plot showing the relationship between estimated \textit{2025} latrine accessibility scores and the reported number of people per functional latrine across 33 camps in Cox’s Bazar District \cite{UNHCR24_WASH}. The red line represents the regression trend, with a strong negative Spearman’s rank correlation ($\rho$ = -0.709, $\rho$ $<$ 0.001). Spearman’s correlation was employed as the variables did not satisfy the normality assumptions required for Pearson’s correlation.}
\label{supply:s4_spearman}
\end{figure*}

\begin{figure*}[t!]
\centering
\includegraphics[width=14cm]{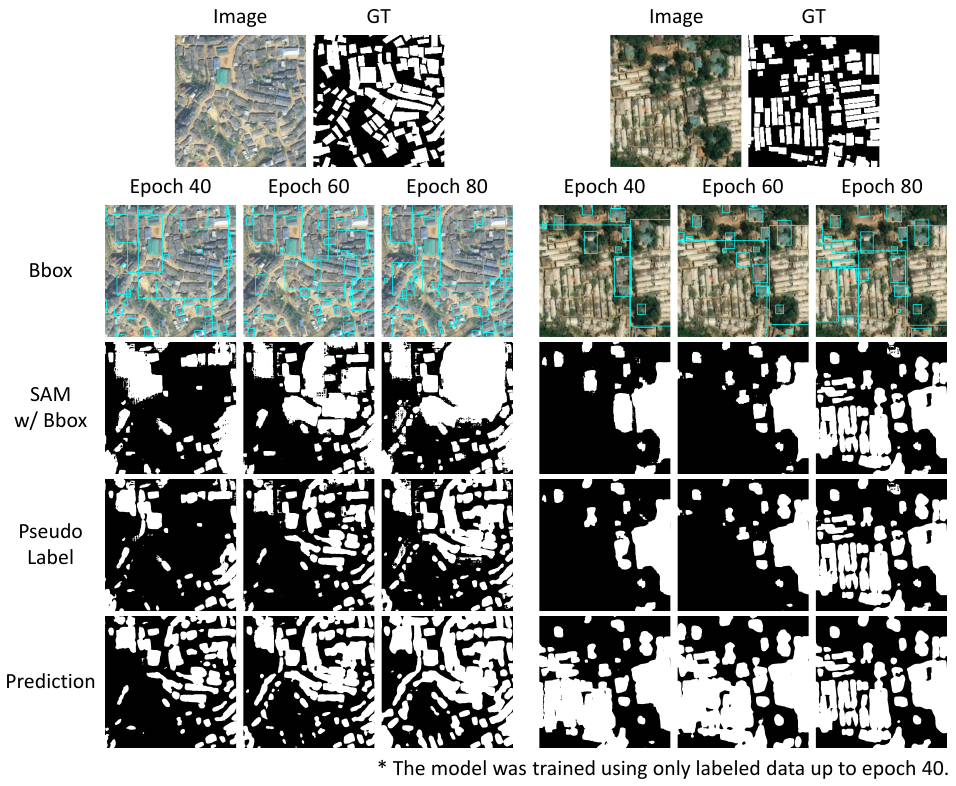}
\caption{\textbf{Progressive Refinement of SAM-guided Pseudo-Labels in High-density Camp Scenarios.} Panels display qualitative improvement for UAV (left) and satellite (right) data across training epochs. The satellite image is from the ESRI World Imagery Wayback; the UAV image is from the International Organization for Migration (IOM).}
\label{supply:s5_progressive_refinement}
\end{figure*}

\end{appendices}

\end{document}